\documentclass{article}

\usepackage{arxiv}

\usepackage[T1]{fontenc}
\usepackage[utf8x]{inputenc}
\usepackage{csquotes}
\usepackage[hidelinks]{hyperref}       
\usepackage{url}            
\usepackage{booktabs}       
\usepackage{amsfonts}       
\usepackage{nicefrac}       
\usepackage{microtype}      
\usepackage{graphicx}
\usepackage{natbib}
\usepackage{doi}

\title{Does Burrows' Delta really confirm that Rowling and Galbraith are the same author?}

\author{\href{https://orcid.org/0000-0002-9099-0436}{\includegraphics[scale=0.06]{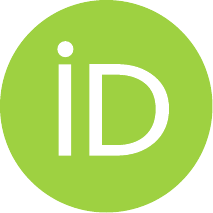}\hspace{1mm}Boris Orekhov} \\
	School of Linguistics\\
	HSE University,\\
	Institute of Russian Literature (Pushkin House)\\
	Russian Academy of Sciences \\
	\texttt{borekhov@hse.ru} 
}

\renewcommand{\shorttitle}{Does Burrows' Delta really confirm that Rowling and Galbraith are the same author?}

\hypersetup{
pdftitle={Does Burrows' Delta really confirm that Rowling and Galbraith are the same author?},
pdfsubject={cs.LG, cs.CL},
pdfauthor={Boris Orekhov},
pdfkeywords={stylometry, rowling case, burrows' delta, stylo},
}

\begin{document}
\maketitle

\begin{abstract}
	The stylo package includes a frequency table that can be used to calculate distances between texts and thus independently solve the problem of attribution of The Cuckoo's Calling, a novel that J.K. Rowling said she wrote. However, the set of texts for this table is very vulnerable to criticism. The authors there are not modern, they wrote in a different genre. I set out to test the performance of the method on texts that are more relevant to the research question.
\end{abstract}


\section{Introduction}

In the humanities, it is rarely possible to resort to proof. Humanities are not built on the formulation of hypotheses and their proof or refutation. It is a field where different ways of describing its material (e.g., artistic culture) compete \citep{harpham2013finding}. Therefore, the question of text authorship is so important for humanists; it remains one of the few questions in the humanities that can be formulated as falsifiable and sometimes verifiable hypotheses. This is an area where humanists find themselves in a situation very similar to that in which representatives of the sciences usually exist.

Consequently, this is the rhetorical resource that humanists can use in the struggle for resources in science and for symbolic capital in the scientific field. The search for the possibility of quantitatively (using calculations) solving the authorship question (the attribution method) has been ongoing since the late 19th century. However, only in the 21st century have sufficiently reliable and verifiable methods for determining authorship on different materials emerged. The state-of-the-art among them in the field of digital humanities has become the Delta method, introduced by John Burrows in 2002 \citep{burrows2002delta}. A significant role in its popularization was played by the remarkable \texttt{stylo} package for the R language \citep{Eder2016Stylometry}. This package provides extensive capabilities for stylometric analysis, flexible parameter selection settings, and a GUI for those researchers who do not know programming languages. The package code contains an acknowledgment: "We have invested a lot of time and effort in creating 'stylo'." \href{https://cran.r-project.org/web/packages/stylo/citation.html}{weblink} And we have no reason to doubt this; the package is indeed made with high quality, considering different situations when working with different languages and cases.

In the extensive functionality of the package, there is the function \texttt{data("galbraith")}, which allows access to word frequency data in a dataset described as follows:

\begin{displayquote}
This dataset contains a table (matrix) of relative frequencies of 3000 most frequent words retrieved from 26 books by 5 authors, including the novel "Cuckoo's Calling" by a mysterious Robert Galbraith that turned out to be J.K. Rowling. The remaining authors are as follows: Harlan Coben ("Deal Breaker", "Drop Shot", "Fade Away", "One False Move", "Gone for Good", "No Second Chance", "Tell No One"), C.S. Lewis ("The Last Battle", "Prince Caspian: The Return to Narnia", "The Silver Chair", "The Horse and His Boy", "The Lion, the Witch and the Wardrobe", "The Magician's Nephew", "The Voyage of the Dawn Treader"), J.K. Rowling ("The Casual Vacancy", "Harry Potter and the Chamber of Secrets", "Harry Potter and the Goblet of Fire", "Harry Potter and the Deathly Hallows", "Harry Potter and the Order of the Phoenix", "Harry Potter and the Half-Blood Prince", "Harry Potter and the Prisoner of Azkaban", "Harry Potter and the Philosopher's Stone"), and J.R.R. Tolkien ("The Fellowship of the Ring", "The Two Towers", "The Return of the King").\footnote{URL: \href{https://rdrr.io/cran/stylo/man/galbraith.html}{https://rdrr.io/cran/stylo/man/galbraith.html}}
\end{displayquote}

The story of the so-called "mysterious" writer Robert Galbraith unfolded as follows. In 2013, a novel titled The Cuckoo's Calling was published under this name. In July of that same year, the Sunday Times wrote \citep{mostrous2013jk} that they had received confirmation from Rowling that the novel was hers. At the same time, the newspaper had the results of a stylometric analysis performed by Peter Millican, who teaches philosophy and computing at Oxford University, and Patrick Juola, a computer science professor at Duquesne University in Pittsburgh. It is believed that these results prompted Rowling to admit the authorship \citep{diep2013how}. Patrick Juola provided a detailed explanation of how these results were obtained \citep{juola2013rowling}. Juola obtained e-text copies of the books involved in the study and used the JGAAP (Java Graphical Authorship Attribution Program) \citep{juola2009jgaap} for preprocessing and analytical operations. Juola conducted four tests, comparing in chunks of 1,000 lines the length of individual words, their frequencies (for the 100 most common words), pairs of words that often appeared together, and frequencies of letter 4-grams. None of the tests conducted by Juola included Burrows' Delta, although the focus on the 100 most frequent words vaguely resembles this method. The study did not yield conclusive results, although J.K. Rowling emerged as the most likely candidate for the authorship of The Cuckoo's Calling. This suspicion formed the basis for the newspaper's inquiry, which was later confirmed.

\begin{figure}
	\centering
	\includegraphics[width=0.7\textwidth]{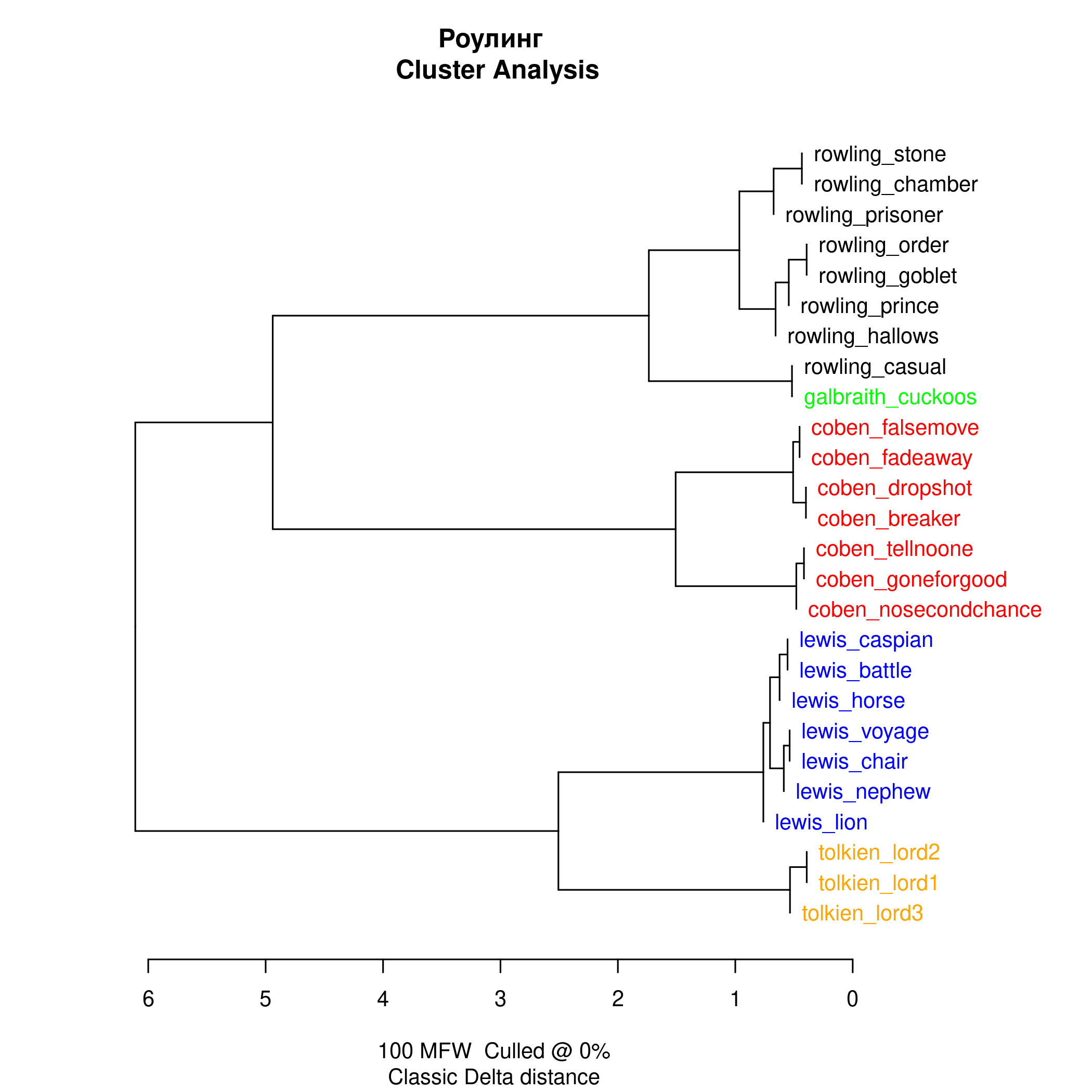}
	\caption{Cluster analysis of the data included in stylo.}
	\label{fig:fig1}
\end{figure}

Thus, the hypothesis regarding the authorship of The Cuckoo's Calling became verifiable and was ultimately confirmed. This is undoubtedly a success story for stylometry and the humanities in general. However, Burrows' Delta did not play a role in this story. Nevertheless, the \texttt{stylo} package uses The Rowling Case to demonstrate the capabilities of various metrics, including Burrows' Delta. The calculations on the built-in package data yield a confident result and clearly show (better than Juola's results) that Rowling was the author of The Cuckoo's Calling. The result is easily obtained using the \texttt{stylo} package, which generates a visualization, as shown in Figure \ref{fig:fig1}.

Code to obtain this visualization:

\begin{verbatim}
library(stylo)
data(galbraith)
setwd('some/folder')
stylo(frequencies = galbraith, gui = FALSE, write.png = TRUE)
\end{verbatim}

However, a closer look reveals many more differences between the study conducted by Juola and the dataset offered in \texttt{stylo}. It’s not just about the methods used but also the texts provided as distractors.

First, the creators of \texttt{stylo} chose texts by Harlan Coben, C.S. Lewis, and J.R.R. Tolkien as contrastive material. Of these three, only Harlan Coben is an author of detective works, and The Cuckoo's Calling is a detective novel. We know that Delta is sensitive to genre differences. Therefore, it is hard to assume that the fantasy works by C.S. Lewis and J.R.R. Tolkien would resemble a detective novel. Including texts that are obviously dissimilar to the investigated text in the dataset makes the analysis result predictable and unjustifiably simplifies the task for the authorship attribution method.

Second, we know that Delta is sensitive to changes over time. The intertextual distances calculated using Delta can even differentiate between early and late works of the same author (see the Nabokov case \citep{orekhov2021text}). C.S. Lewis and J.R.R. Tolkien are writers from another era, having passed away several decades before Rowling published her first novel. It is hard to expect that the method will show similarities to their works under such simplified conditions. Harlan Coben is a contemporary of Rowling, but he is American, representing a different linguistic culture, and from the outset, he is just as distant from Rowling as C.S. Lewis and J.R.R. Tolkien.

Therefore, in Juola's study, the distractors were correctly chosen: three British women who specialize in crime fiction, Ruth Rendell's The St. Zita Society, P.D. James' The Private Patient, and Val McDermid's The Wire in the Blood. Juola did not play favorites and honestly tried to answer whether the styles of Rowling and Galbraith's novels are similar or not. Overall, it can be stated that the \texttt{stylo} dataset proposed for authorship attribution is very peculiar and does not meet the minimum requirements of scientific standards.

In this article, I will conduct a comparative stylometric study on The Rowling Case using the \texttt{stylo} package with texts more appropriate than those suggested in the \texttt{data("galbraith")} dataset. I will try to formulate what the answer would be if the Sunday Times had approached a scholar who uses \texttt{stylo} and Burrows' Delta instead of JGAAP for their work.

\section{Data}

The novels used in Juola's study, although correctly selected, were still random, and within the same selection principles (contemporary crime fiction writers), any other works can be included in the research corpus.

Additionally, Rowling's original Harry Potter works represent a modern variant of the fantasy genre, different from the fantasy of C.S. Lewis and J.R.R. Tolkien. From this perspective, it is appropriate to compare Rowling's novels written under her own name with Galbraith's novel against the backdrop of contemporary fantasy literature.

The creators of \texttt{stylo} write: "The novels represented by this dataset are protected by copyright. For that reason, it was not possible to provide the actual texts. Instead, the frequencies of the most frequent words are obtained – and those can be freely distributed." I followed the same path and published the data generated by \texttt{stylo} when working with the texts, making it freely available, as copyright considerations also prevent me from publishing the original texts \citep{orekhov2024rowling}.

Since Rowling's genres (under her pseudonym and real name) are fantasy (the Harry Potter books) and detective fiction (The Cuckoo's Calling), I decided to compile two contrastive corpora from contemporary texts in these genres.

Detective novels included in the corpus:

\begin{itemize}
  \item The Secret Place (2014) by Tana French 
  \item The Trespasser (2016) by Tana French 
  \item The Disappearance of Adèle Bedeau (2014) by Graeme Macrae Burnet
  \item His Bloody Project: Documents relating to the case of Roderick Macrae (2015) by Graeme Macrae Burnet
  \item The Seven Deaths of Evelyn Hardcastle (2018) by Stuart Turton
  \item The Devil and the Dark Water (2020) by Stuart Turton
\end{itemize}

The total volume of this corpus (excluding Rowling's texts) is 815,749 tokens.

Novels included in the fantasy corpus:

\begin{itemize}
  \item A Blackbird in Silver (1986) by Freda Warrington
  \item The Dark Arts of Blood (2015) by Freda Warrington
  \item Mistborn: The Final Empire (2006) by Brandon Sanderson
  \item Mistborn: The Well of Ascension (2007) by Brandon Sanderson
  \item Mistborn: The Hero of Ages (2008) by Brandon Sanderson
  \item The Amulet of Samarkand (2003) by Jonathan Stroud
  \item The Golem's Eye (2004) by Jonathan Stroud
  \item Ptolemy's Gate (2005) by Jonathan Stroud
\end{itemize}

The total volume of this corpus is 1,327,578 tokens.

These texts are protected by copyright, so I cannot make them publicly available. However, the word frequencies necessary for stylometric analysis can be published and can be found here: \citep{orekhov2024rowling}.

\section{Results}

\begin{figure}
	\centering
	\includegraphics[width=0.7\textwidth]{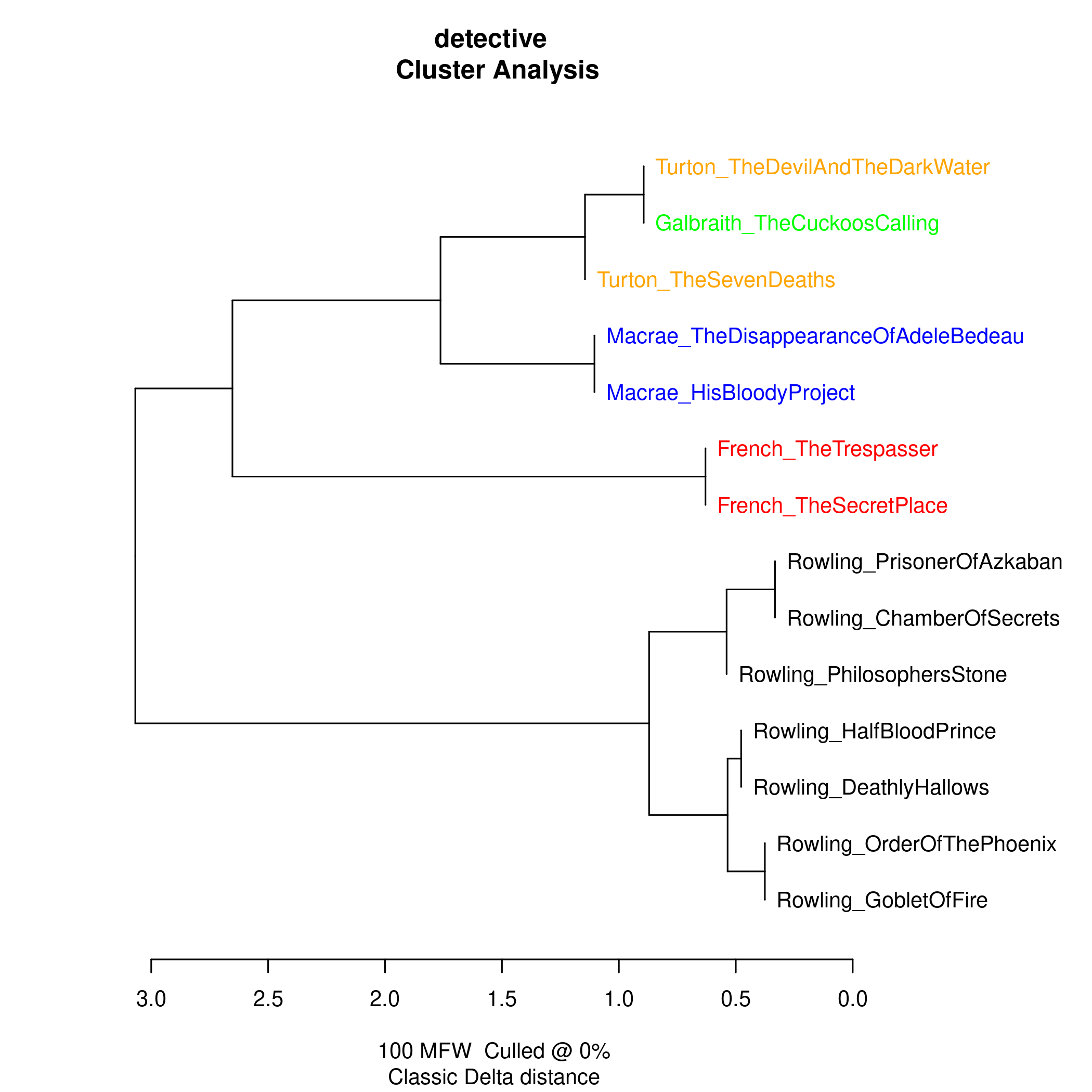}
	\caption{Cluster analysis of the detective data.}
	\label{fig:fig2}
\end{figure}

Using \texttt{stylo}, I analyzed the corpus that includes detective novels and Rowling's books. The result can be seen in the \texttt{stylo} built-in visualization in Fig. \ref{fig:fig2}.

The distances between the novels are given in Table \ref{table1}. The full data available at \citep{orekhov2024rowling}.

\begin{table}[!ht]
    \centering
    \begin{tabular}{llll}
    \hline
        ~ & French\_TheSecretPlace & French\_TheTrespasser & Galbraith\_TheCuckoosCalling  \\ \hline
        French\_TheSecretPlace & 0 & 0.6296 & 1.2243  \\ 
        French\_TheTrespasser & 0.6296 & 0 & 1.2695  \\ 
        Galbraith\_TheCuckoosCalling & 1.2243 & 1.2695 & 0\\ 
        Macrae\_HisBloodyProject & 2.0114 & 2.0217 & 1.4310 \\ 
        Macrae\_TheDisappearanceOfAdeleBedeau & 1.9456 & 1.9379 & 1.0781 \\ 
        Rowling\_ChamberOfSecrets & 1.2352 & 1.3937 & 0.8599  \\ 
        Rowling\_DeathlyHallows & 1.3717 & 1.4649 & 0.8813  \\ 
        Rowling\_GobletOfFire & 1.3829 & 1.5502 & 0.8787  \\ 
        Rowling\_HalfBloodPrince & 1.2686 & 1.3657 & 0.8320 \\ 
        Rowling\_OrderOfThePhoenix & 1.2195 & 1.3899 & 0.8256  \\ 
        Rowling\_PhilosophersStone & 1.1041 & 1.2019 & 0.9396  \\ 
        Rowling\_PrisonerOfAzkaban & 1.2453 & 1.3990 & 0.8589  \\ 
        Turton\_TheDevilAndTheDarkWater & 1.3149 & 1.3788 & 0.8938  \\ 
        Turton\_TheSevenDeaths & 1.4243 & 1.2358 & 1.2239  \\ \hline
    \end{tabular}
    \label{table1}
    \caption{Detective novels, intertextual distanses}
\end{table}

\begin{figure}
	\centering
	\includegraphics[width=0.7\textwidth]{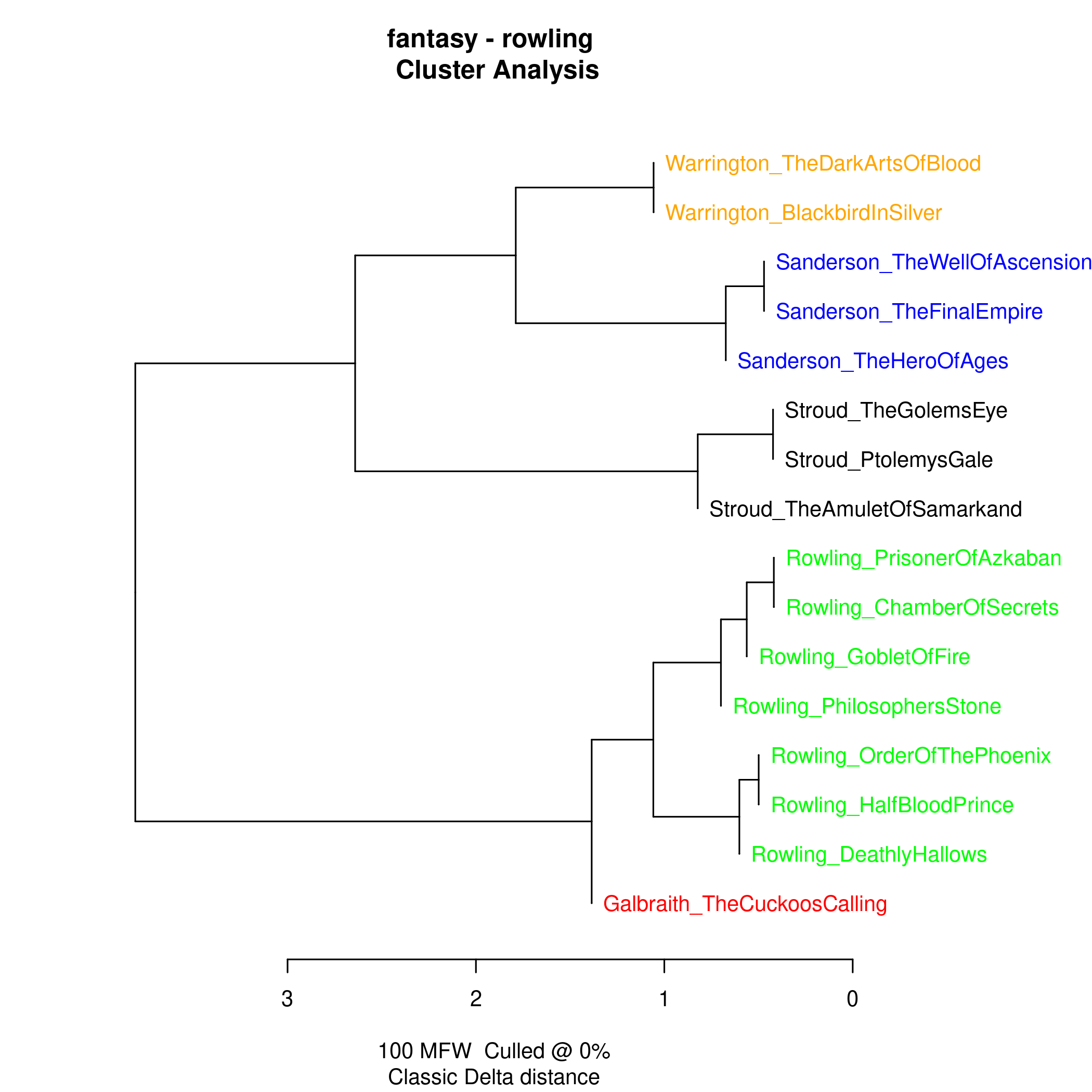}
	\caption{Cluster analysis of the fantasy data.}
	\label{fig:fig3}
\end{figure}

A similar analysis was conducted for the fantasy corpus, with visualization in Fig. \ref{fig:fig3} and in Table \ref{table2}.

\begin{table}[!ht]
    \centering
    \begin{tabular}{llll}
    \hline
        ~ & Galbraith\_TheCuckoosCalling & Rowling\_ChamberOfSecrets & Rowling\_DeathlyHallows \\ \hline
        Galbraith\_TheCuckoosCalling & 0 & 1.0606 & 1.0608  \\ 
        Rowling\_ChamberOfSecrets & 1.0606 & 0 & 0.6734  \\ 
        Rowling\_DeathlyHallows & 1.0608 & 0.6734 & 0  \\ 
        Rowling\_GobletOfFire & 1.1117 & 0.5570 & 0.6285  \\ 
        Rowling\_HalfBloodPrince & 1.0009 & 0.7866 & 0.6093 \\ 
        Rowling\_OrderOfThePhoenix & 1.0346 & 0.6140 & 0.5427 \\ 
        Rowling\_PhilosophersStone & 1.1147 & 0.5706 & 0.8144  \\ 
        Rowling\_PrisonerOfAzkaban & 1.0786 & 0.4186 & 0.6681  \\ 
        Sanderson\_TheFinalEmpire & 1.2168 & 1.2883 & 1.2350  \\ 
        Sanderson\_TheHeroOfAges & 1.2384 & 1.2650 & 1.1139  \\ 
        Sanderson\_TheWellOfAscension & 1.2136 & 1.2477 & 1.1055  \\ 
        Stroud\_PtolemysGale & 1.1982 & 1.2922 & 1.2532  \\ 
        Stroud\_TheAmuletOfSamarkand & 1.2877 & 1.3594 & 1.3954  \\ 
        Stroud\_TheGolemsEye & 1.2947 & 1.3212 & 1.2916  \\ 
        Warrington\_BlackbirdInSilver & 1.4133 & 1.3340 & 1.0839  \\ 
        Warrington\_TheDarkArtsOfBlood & 1.3899 & 1.4136 & 1.3283  \\ \hline
    \end{tabular}
    \caption{Fantasy novels, intertextual distances}
    \label{table2}
\end{table}

\begin{figure}
	\centering
	\includegraphics[width=0.7\textwidth]{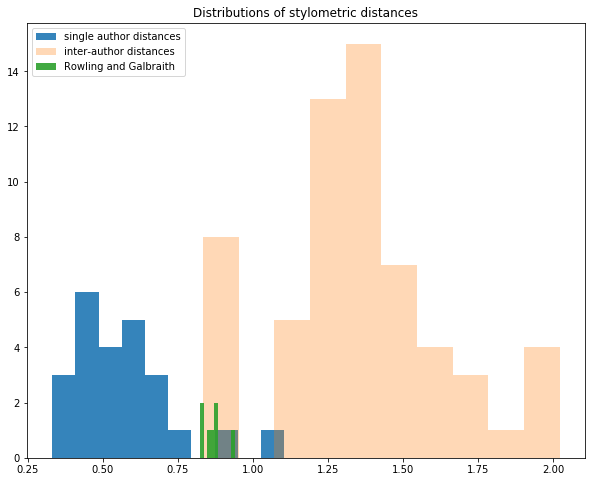}
	\caption{Distribution of stylometric distances in detective fiction corpus.}
	\label{fig:fig4}
\end{figure}

\begin{figure}
	\centering
	\includegraphics[width=0.7\textwidth]{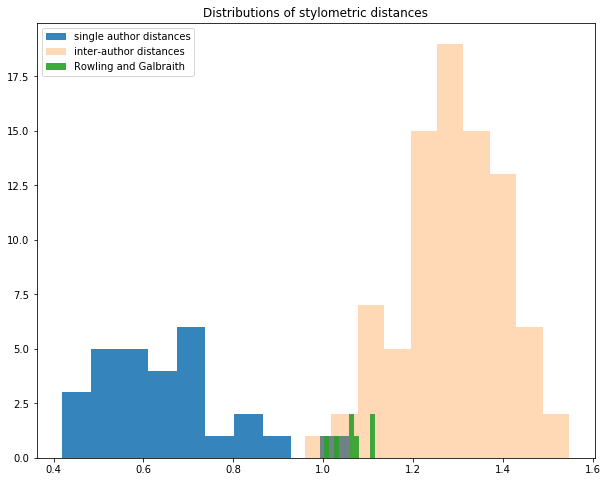}
	\caption{Distribution of stylometric distances in fantasy fiction corpus.}
	\label{fig:fig5}
\end{figure}

You can reproduce this result by using the data from this repository \citep{orekhov2024rowling} and loading them into R with the following code:

\begin{verbatim}
setwd('folder/with/data/unpacked')
stylo()
\end{verbatim}

My co-author and I previously suggested this method of visualizing intertextual distances \citep{skorinkin2023hacking}, and this idea was well received by the academic community \citep{schoch2023dear}. Let's look at the results using this summarization method, see Figs. \ref{fig:fig4} and \ref{fig:fig5}. The code for generating this visualization is available in the publication \citep{orekhov2024rowling}.

\begin{figure}
	\centering
	\includegraphics[width=1\textwidth]{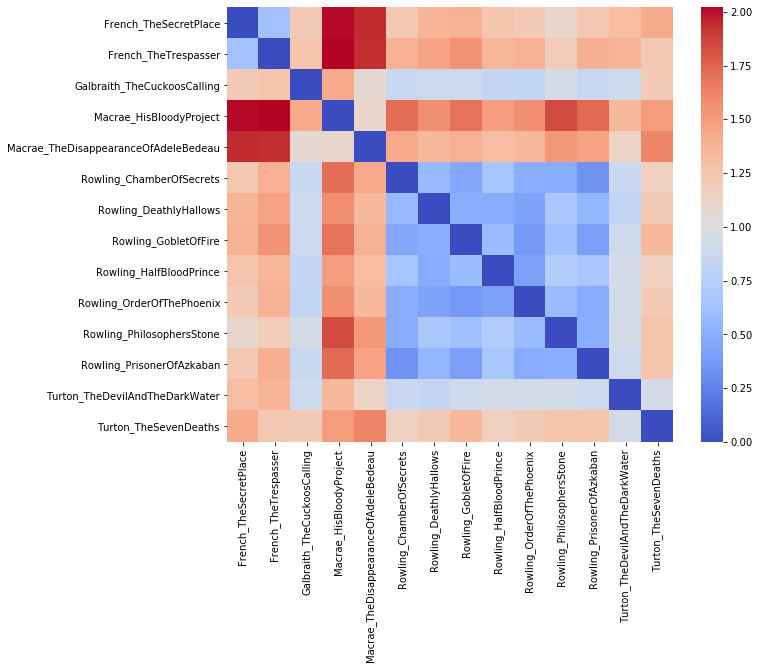}
	\caption{Heatmap of the intertextual distances between detective texts.}
	\label{fig:fig6}
\end{figure}

Fig. \ref{fig:fig6} shows the heatmap of intertextual distances in the contemporary detective fiction corpus.

Furthermore, \texttt{stylo} includes a function for authorship attribution using the imposters algorithm. On the built-in Galbraith data, this function always shows a definitive result. Let's see what the result will be with our data. Here is the code for this operation:

\begin{verbatim}
detective <- read.csv("table_with_frequencies.txt", sep=' ')
df.detective <- t(detective)
imposters(df.detective)
\end{verbatim}

Here is the output for this code using the detective data:

\begin{verbatim}
No candidate set specified; testing the following classes (one at a time):
  French   Macrae   Rowling   Turton
 
Testing a given candidate against imposters...

French 	 0.08
Macrae 	 0.69
Rowling 	 1
Turton 	 0
 French  Macrae Rowling  Turton 
   0.08    0.69    1.00    0.00
\end{verbatim}

Here is the output using the fantasy data:

\begin{verbatim}
No candidate set specified; testing the following classes (one at a time):
  Rowling   Sanderson   Stroud   Warrington
 

Testing a given candidate against imposters...

Rowling 	 1
Sanderson 	 0
Stroud 	 0.02
Warrington 	 0.05
   Rowling  Sanderson     Stroud Warrington 
      1.00       0.00       0.02       0.05 
\end{verbatim}

\section{Discussion}

As we know from related materials: "The decision to choose a male pseudonym was driven by a desire to 'take my writing persona as far away as possible from me', Rowling said" \citep{bury2013jk}. However, according to the study \citep{skorinkin2023hacking}, there is a significant difference in the results of stylometric research depending on whether an author simply adopts a pseudonym or creates a separate persona, an individual "voice." In this case, choosing a pseudonym was not a mere formality; it involved creating a fully-fledged biographical alternative author.

When relying on classic methods of representing results, Delta confirms the hypothesis of Rowling's authorship against the backdrop of contemporary fantasy and provides an uncertain answer against the backdrop of detective novels. This is likely due to the individuality of each fantasy world and the unique means of depicting it in each specific case. At the same time, the detective genre is formulaic \citep{cawelt2014adventure}, and different authors' books share many common elements.

However, more adequate forms of visualizing results (Fig. \ref{fig:fig6}) show that Rowling remains the most likely candidate for the authorship of Galbraith's novel. Nevertheless, this case is borderline, as the intertextual distances between Rowling and Galbraith are in the pink zone of inter-author distances in both Fig. \ref{fig:fig4} and Fig. \ref{fig:fig5}. This effect is even more pronounced in the case of fantasy.

Simultaneously, the data from the impostor algorithm, which was integrated into \texttt{stylo} in 2018 (\href{https://computationalstylistics.github.io/docs/imposters}{weblink}), provides a strong argument in favor of Rowling's authorship against the backdrop of authors writing in any genre. However, in the case of detective novels, Graeme Macrae Burnet shows suspiciously high figures, suggesting he could also be a potential author of The Cuckoo's Calling.

\section{Conclusion}

As is often the case in stylometry, we did not find definitive answers. Delta demonstrated its effectiveness in some forms and yielded uncertain results in others. However, a positive outcome is that Delta did not fail in new tests on more suitable texts.

The most promising result came from the impostor detection method, which shows correct results on any texts. It seems necessary to test this method on various languages and authors, for which intertextual distances obtained using Delta can be employed.

\section*{Acknowledgements}
I am grateful to my colleague and co-author Daniil Skorinkin for fruitful discussions on many details of this work, bibliographic suggestions, and assistance with the code.

\bibliographystyle{unsrtnat}
\bibliography{references} 

\end{document}